\documentclass[letterpaper, 10 pt, conference]{ieeeconf}  

\IEEEoverridecommandlockouts                              

\usepackage{graphicx} 
\usepackage{xcolor}
\usepackage{url}
\usepackage{booktabs}
\usepackage{listings}
\usepackage{hyphenat}
\usepackage[english]{babel}
\usepackage{tabularx}
\usepackage{array}
\usepackage{makecell}
\usepackage[table]{xcolor}

\title{\LARGE \bf
Can Robot Narrative Complexity Influence Human Production and Engagement in Collaborative  Storytelling?}

\author{Carmela Calabrese$^{1,\ast}$, Maria Lombardi$^{1}$, Caterina Foglino$^{2}$, \\ Marco Nasato$^{1}$, Federico Conti$^{3}$, Agnieska Wykowska$^{2, \S}$, Lorenzo Natale$^{1, \S}$
\thanks{\scriptsize{$\ddagger$Corresponding author: {\tt carmela.calabrese@iit.it}}}%
\thanks{\scriptsize{$^{\S}$Shared last authorship}}%
\thanks{\scriptsize{$^{1}$C.C., M.M., M.N., and L.N. are with Humanoid Sensing and Perception, Italian Institute of Technology (IIT), Genoa, 16163, Italy; $^{2}$C.F. and A.W. are with Social Cognition in Human-Robot Interaction, IIT, Genoa, 16163, Italy; $^{3}$F.C. is with Information and Communication Technology Directorate, IIT, Genoa, 16163, Italy}} 
\thanks{\scriptsize{This work has been carried out within the Brain and Machines Flagship Programme of IIT and received funding from the project Fit for Medical Robotics (Fit4MedRob) - PNRR MUR Cod. PNC0000007 - CUP: B53C22006960001 and from the European Research Council (ERC) under the European Union’s Horizon Europe research and innovation programme (Grant title ``RONIN - Robot Training Independence'', agreement No. ERC-2023-POC-101155938.}}
}

\begin{document}

\maketitle
\thispagestyle{empty}
\pagestyle{empty}

\begin{abstract}
Large language models have enabled robots to engage in adaptive storytelling interactions, opening new possibilities for their use in cognitive rehabilitation. However, limited attention has been devoted to understanding how the narrative behavior of a robot influences the contribution of the human partner. This work investigates whether controlling the narrative complexity of a humanoid modulates human narrative production during collaborative storytelling. We extend a previously developed architecture by introducing an automatic narrative-complexity analyzer based on the Monitoring Indicators of Scholarly Language (MISL) framework and by tuning the generation pipeline to produce two different narrative behaviours. Experiments in a within-subject study show that participants produced significantly more words and stories with higher microstructure scores in the complex condition, aligning to robot's narrative strategy. Participants also rated stories in the Complex condition as more likable and coherent and expressed preference for this interaction mode. The results suggest that the strategy adopted by a storytelling partner can influence human narrative behavior and attunement to the robot. These findings provide a first step toward adaptive narrative human--robot interaction paradigms for cognitive stimulation and, in the longer term, assessment and rehabilitation.
\end{abstract}

\section{INTRODUCTION}

Socially interactive robots are designed not only to assist users physically, but also to engage them in cognitive activities \cite{tamantini2023integrating}. Their behavior shapes both the structure and perceived social nature of the interaction, influencing the user's experience and responses.

Collaborative storytelling provides a suitable paradigm for investigating this influence in human---robot interaction (HRI). It is inherently sequential and collaborative: each partner must interpret the evolving narrative context, maintain temporal and causal coherence, and introduce new characters, events, or intentions through turn-taking and shared visual stimuli to jointly construct a story.

Storytelling is cognitively and linguistically demanding. Producing a coherent narrative requires organizing events, representing characters' goals and internal states, maintaining information across turns, and expressing these elements through appropriate lexical and grammatical structures. Consequently, narrative production has been investigated as an indicator of mental functioning and as a tool for cognitive rehabilitation \cite{choi2025narrative}, \cite{dosi2025case}. These properties make collaborative storytelling particularly relevant to socially assistive robotics (SAR), where robots can at once foster engagement, elicit narrative production, and adapt to the user's responses.

Beyond assisting users during physical rehabilitation, socially assistive robots are increasingly being employed as interactive partners capable of delivering personalized cognitive stimulation and adapting their behavior according to individual abilities \cite{ghiglino2023artificial, kubota2022methods,ilyas2022deep}.
Recent studies have also demonstrated the feasibility of using humanoid robots to autonomously administer standardized assessment procedures while maintaining user engagement \cite{calabrese2025socially}. These advances suggest that socially assistive robots may become not only therapeutic tools \cite{ghiglino2025reimagining}, but also adaptive platforms to monitor users' abilities through natural interaction.

\begin{figure}[t]
    \centering
    \includegraphics[width=1\columnwidth]{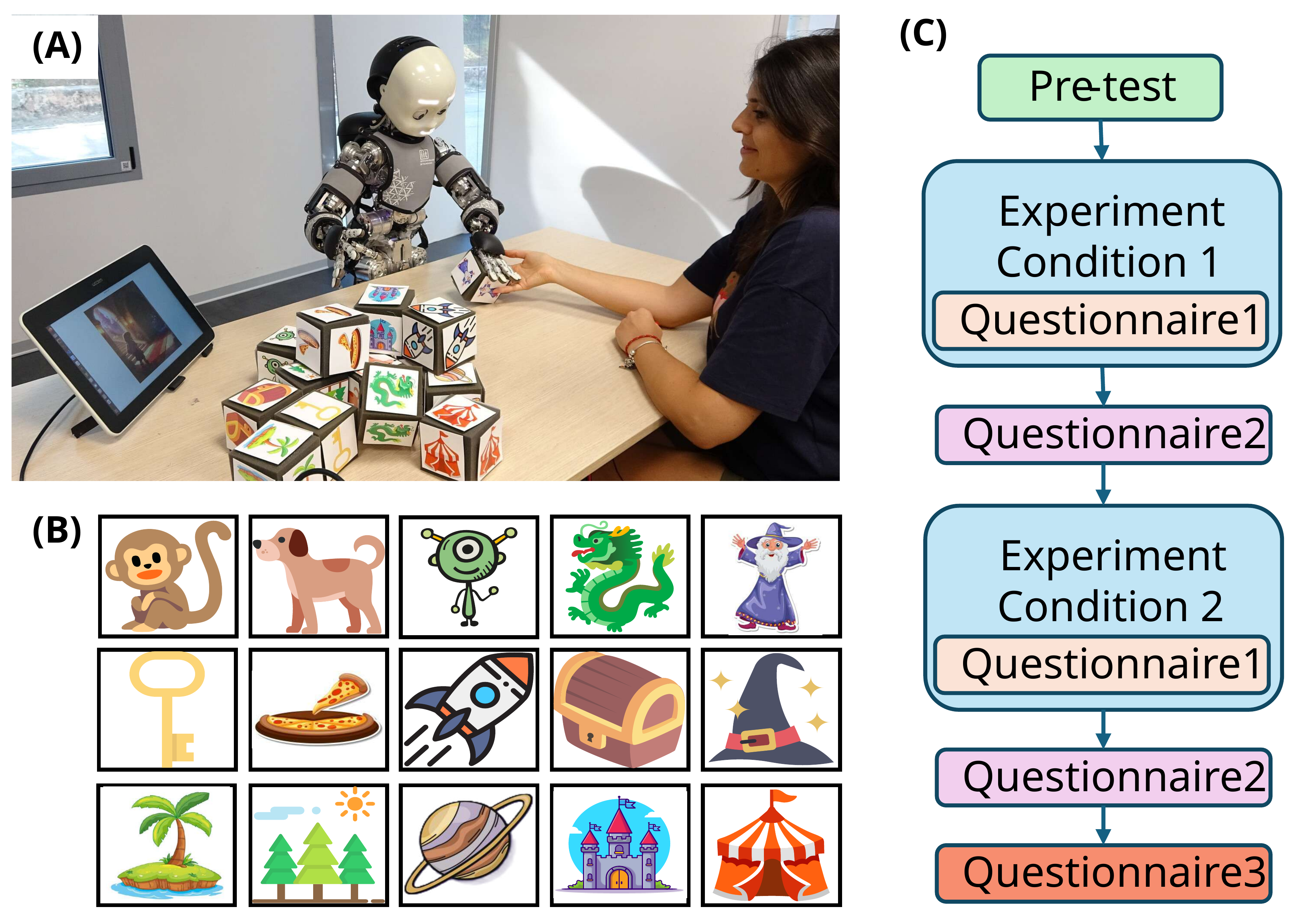}
    \caption{\textbf{Collaborative storytelling task and experimental procedure.} (A) HRI setup with the participant interacting with iCub through cubes and a shared visual display. (B) Set of visual stickers used as narrative stimuli. (C) Experimental procedure, including the pre-test, the two storytelling conditions, the questionnaires administered after each trial and condition, and the final post-experiment comparison.}
    \label{fig:setup}
\vspace{-3mm}
\end{figure}

Recent advances in large language models (LLMs) expanded the communicative capabilities of interactive robots, enabling context-aware narrative generation, multimodal interaction, and open-ended storytelling. Nevertheless, existing research has primarily focused on improving the quality and personalization of AI-generated stories \cite{sun2025drama, valentini2023automatic}, while less attention has been devoted to a central HRI question: how does the communicative behavior of an embodied artificial agent influence the human partner's contribution?

A robot may adopt different storytelling strategies. Richer narrative turns can provide linguistic and conceptual scaffolding, whereas sparser contributions may leave greater narrative space for the user. Narrative complexity can therefore be viewed not only as a property of generated content, but also as a tunable parameter that shapes human participation.

Our previous framework enabled participants and the humanoid robot iCub to jointly construct stories from participant-selected visual stimuli \cite{lombardi2025would}. Although usability and enjoyment were positively evaluated, some participants found the stories insufficiently rich or engaging. This motivated us to investigate whether increasing the robot’s narrative complexity changes human continuations and the perceived quality of the interaction.

To address this question, we extended the original architecture with an automatic narrative analyzer based on the Monitoring Indicators of Scholarly Language (MISL) framework \cite{gillam2017monitoring}, which evaluates story organization and linguistic complexity. The analyzer was validated against a consensus reference derived from independent human raters and subsequently used in an iterative prompt calibration process, resulting in two distinct robot behaviors: one producing shorter, structurally simpler narrative turns and the other generating richer and more complex descriptions.

The resulting system was evaluated in a within-subject HRI study in which participants jointly created stories with iCub under both conditions. Human narrative production was assessed in terms of word count, linguistic microstructure, and macrostructural development during the participant's contribution. Participants also evaluated the interaction by rating the likability and coherence of each jointly generated story, together with usability, engagement, preference, and perceived contribution. Accordingly, this study addresses the following research questions:

\begin{itemize}
    \item \textbf{RQ1:} Are human verbal productivity and linguistic narrative complexity
    influenced by the narrative strategy adopted by the humanoid robot?

    \item \textbf{RQ2:} How does the narrative developed with the human compare with the complexity that the robot would autonomously produce from the same interaction context?

    \item \textbf{RQ3:} Does a more complex storytelling strategy improve perceived story quality and interaction preference without negatively affecting usability or engagement?

    \item \textbf{RQ4}: Are individual cognitive abilities associated with collaborative storytelling performance?
\end{itemize}

The main contributions of this work are:

\begin{itemize}
    \item an automatic MISL-based analyzer evaluated against a human consensus reference and used to calibrate distinct robot narrative behaviors;

    \item a complexity-controlled narrative interaction architecture for collaborative storytelling with iCub;

    \item a within-subject HRI study investigating whether the narrative complexity adopted by iCub elicits adaptation of its human partner and evaluation of the interaction.
\end{itemize}

The findings provide a basis for future SARs capable of dynamically regulating their narrative behavior according to the user's capabilities and interaction goals.

\section{RELATED WORKS}

\subsection{Foundation models for narrative architectures}

LLMs have significantly expanded the capabilities of interactive storytelling systems, enabling coherent story generation, adaptive plot development, and natural language interaction between users and artificial agents. Recent work has explored their integration into narrative architectures from multiple perspectives.

\textit{ToyTeller} \cite{chung2025toyteller} combines LLM-based generation with tangible character symbols to support children's visual storytelling, while \textit{Drama Llama} \cite{sun2025drama} employs a storylet-based architecture to dynamically adapt plot progression according to user actions. Similarly, \textit{StoryVerse} \cite{wang2024storyverse} leverages LLM-driven character simulation and narrative planning to support collaborative story creation.

Other studies have focused on long-form story generation. \textit{SEED-Story} \cite{yang2025seed} combines visual and textual information to maintain coherence over extended narratives, while Valentini \textit{et al.} \cite{valentini2023automatic} demonstrated the potential of foundation models to adapt children's stories to different audiences.

Overall, existing work has primarily aimed to improve the quality, coherence, and personalization of AI-generated narratives. In contrast, less attention has been devoted to understanding whether varying the complexity of AI-generated scaffolds can influence human response. Our work addresses this gap by investigating how the storytelling strategy adopted by a robot partner affects participants' verbal productivity and narrative contribution during joint storytelling.

\subsection{Monitoring narrative complexity for assessment and cognitive intervention}
Narrative production has been recognized as a marker of cognitive functioning because it requires the integration of memory, executive functions, and language organization. Consequently, narrative analysis has increasingly been adopted as a diagnostic tool.

Recent studies have shown that narrative macrostructure can discriminate between healthy aging, mild cognitive impairment (MCI), and Alzheimer's disease. Kaltsa \textit{et al.} \cite{kaltsa2024assessing} reported a progressive deterioration of narrative macrostructure across the Alzheimer's disease continuum, while Mohajeri \textit{et al.} \cite{mohajeri2026evaluating} validated story retelling as a reliable approach for identifying individuals with MCI.

Beyond diagnosis, narrative-based activities have also been investigated as therapeutic interventions. Choi \textit{et al.} \cite{choi2025narrative} reported improvements in executive functions following an interactive storytelling game for individuals with MCI. Narrative analysis has also proved valuable in Autism Spectrum Disorder (ASD), where atypical narrative organization has been linked to social communication difficulties and mentalizing abilities \cite{dosi2025case,wimmer2025autistic}.

These findings highlight narrative complexity as both a clinically meaningful assessment measure and a promising target for cognitive intervention. However, little attention has been devoted to understanding how different narrative strategies of the trainer influence human behavior. Our work addresses this gap by investigating whether the complexity of an LLM-generated narrative scaffold modulates human narrative production during collaborative storytelling, with the long-term goal of AI-assisted cognitive assessment and rehabilitation.

\section{SYSTEM ARCHITECTURE}

This work extends the collaborative storytelling architecture presented in~\cite{lombardi2025would}. The original framework enables a participant and the humanoid robot iCub to alternately contribute, in three steps, to a shared story using visual stimuli on cubes (see Fig. \ref{fig:setup}). The original system in~\cite{lombardi2025would} uses an LLM which is prompted to generate stories based on the visual stimuli seen by the robot and the story parts contributed by the human partner. In the present version, we integrated two additional AI-agents. The first agent evaluates narrative complexity according to the MISL framework, while supporting the iterative calibration of the generation prompt toward a fixed complexity profile. The second agent produces illustrations of the evolving story, helping maintain a shared narrative context throughout the interaction.

\begin{figure*}[t]
    \centering
    \includegraphics[width=1\textwidth]{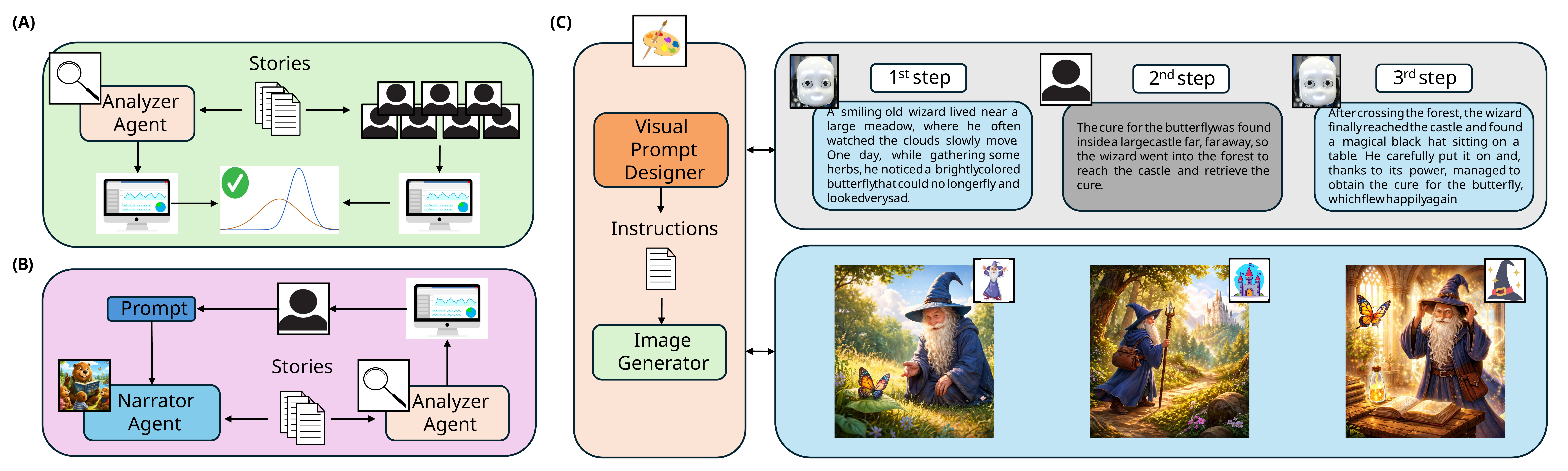}
    \caption{\textbf{AI-based modules.}
(A) Validation of the Analyzer Agent against human MISL ratings.
(B) Human-in-the-loop calibration of the Narrator Agent prompt using  Analyzer Agent's evaluations.
(C) Painter Agent and example of the textual and visual evolution of a 3-steps joint story.}
    \label{fig:agents}
\end{figure*}

\subsection{Analyzer agent: Automatic narrative-complexity analysis}

Narrative complexity was automatically assessed using the MISL framework \cite{gillam2017monitoring} that evaluates a total of twelve items: seven macrostructure items---Character, Setting, Initiating Event (IE), Internal Response (IR), Plan, Attempt, and Consequence---and five microstructure items---Coordinating (CCs) and Subordinating Conjunctions (SCs), Meta-Cognitive Verbs (MCVs), Adverbs, and Elaborated Noun Phrases (ENPs). Each item is scored on a $[0-3]$ scale, yielding maximum macrostructure, microstructure, and global scores of 21, 15, and 36, respectively.

Since manual MISL annotation is time-consuming and requires raters, we implemented a two-stage LLM-based analyzer. First, the model extracts only the narrative evidence associated with each MISL feature present in the text; then, the extracted evidence is scored according to the MISL rubric. Separating extraction from scoring was intended to reduce unsupported inferences and improve evaluation transparency.

To validate the analyzer, we compared the provided evaluations with those of human raters. To this end, 20 stories collected from different children's websites were independently assessed by seven raters (Fig.~\ref{fig:agents}A). Inter-rater reliability for each feature was estimated using Intraclass Correlation Coefficient, ICC(2,k), i.e., a mean-rating, absolute-agreement, two-way random-effects model \cite{koo2016guideline}.
The majority-rule ratings were used as the human consensus reference, producing a mean global MISL score of 25.60 ($SD=4.40$).

The same 20 stories were subsequently evaluated using GPT-4.1. Agreement between GPT-4.1 and the human consensus was quantified using a normalized distance-based agreement score, defined as $A_{\mathrm{norm}}=1-|H-G|/3$, where $H$ and $G$ denote the human-consensus and GPT-4.1 scores, respectively. The denominator corresponds to the maximum possible difference on the 0--3 MISL scale, yielding values from 0 (maximum disagreement) to 1 (perfect agreement). Scores were averaged across stories for each MISL dimension. The automatic analyzer produced a mean global MISL score of 27.50 ($SD=4.65$). Across the twelve MISL dimensions, the mean absolute difference between the human-consensus and GPT-4.1 scores was 0.24 points, while normalized agreement ranged from $70\%$ to $97\%$ (see Table~\ref{table:icc}).

At the subscale level, GPT-4.1 assigned higher macrostructure scores ($M=14.55$, $SD=3.49$) than the human consensus ($M=13.15$, $SD=3.10$). A paired-samples $t$-test confirmed a statistically significant mean difference of 1.40 points,
$t(19)=-3.074$, $p=.006$. Conversely, no statistically significant difference emerged for microstructure scores between GPT-4.1 ($M=12.95$, $SD=1.86$) and the human consensus ($M=12.45$, $SD=2.18$), $t(19)=-1.422$, $p=.17$.

Overall, the analyzer showed close feature-level agreement with the human reference and no evidence of a systematic mean difference in microstructure scoring. However, it exhibited a minimal but statistically significant tendency to overestimate macrostructure scores.

\begin{table}[t]
\centering
\caption{\textbf{Human inter-rater reliability
and agreement between the Analyzer Agent and human
consensus across MISL dimensions.}}
\vspace{-3mm}
\label{table:icc}

\scriptsize
\setlength{\tabcolsep}{3pt}
\renewcommand{\arraystretch}{0.95}

\begin{tabular*}{\columnwidth}{
    @{\extracolsep{\fill}}lcccc@{}
}
\toprule
\textbf{Feature} &
\makecell{\textbf{Human}\\\textbf{mean}} &
\makecell{\textbf{ICC}\\\textbf{(2,k)}} &
\makecell{\textbf{Analyzer}\\\textbf{mean}} &
\makecell{\textbf{Normalized}\\\textbf{agreement}} \\
\midrule

\rowcolor{gray!20}
\multicolumn{5}{l}{\textbf{Macrostructure}} \\
\addlinespace[1pt]

Character    & 1.65 & 94\%   & 1.60 & 95\%   \\
Setting      & 1.70 & 84.3\% & 1.95 & 85\%   \\
IE           & 2.45 & 85.7\% & 2.65 & 86.7\% \\
IR           & 1.90 & 85\%   & 2.05 & 91.7\% \\
Plan         & 1.10 & 57.5\% & 1.60 & 70\%   \\
Attempt      & 2.05 & 79.3\% & 2.20 & 95\%   \\
Consequences & 2.30 & 83.2\% & 2.50 & 86.7\% \\

\midrule
\rowcolor{gray!20}
\multicolumn{5}{l}{\textbf{Microstructure}} \\
\addlinespace[1pt]

CCs     & 2.25 & 74.5\% & 2.40 & 85\% \\
SCs     & 2.60 & 87.6\% & 2.15 & 78.3\% \\
MCV     & 2.75 & 74.5\% & 2.85 & 90\% \\
Adverbs & 1.95 & 79.3\% & 2.55 & 80\% \\
ENPs    & 2.90 & 54.6\% & 3.00 & 97\% \\

\bottomrule
\end{tabular*}
\vspace{-3mm}
\end{table}

\subsection{Complexity-controlled generation}
\label{sec: prompt-tuning}
To obtain robot-generated narratives with a predefined complexity profile, we developed a human-in-the-loop prompt-calibration procedure. Starting from the prompt employed in the previous work \cite{lombardi2025would}, we iteratively revised it and evaluated the resulting narratives using the automatic MISL analyzer described previously.

\begin{figure}[ht]
   \centering
          \includegraphics[width=0.9\columnwidth]{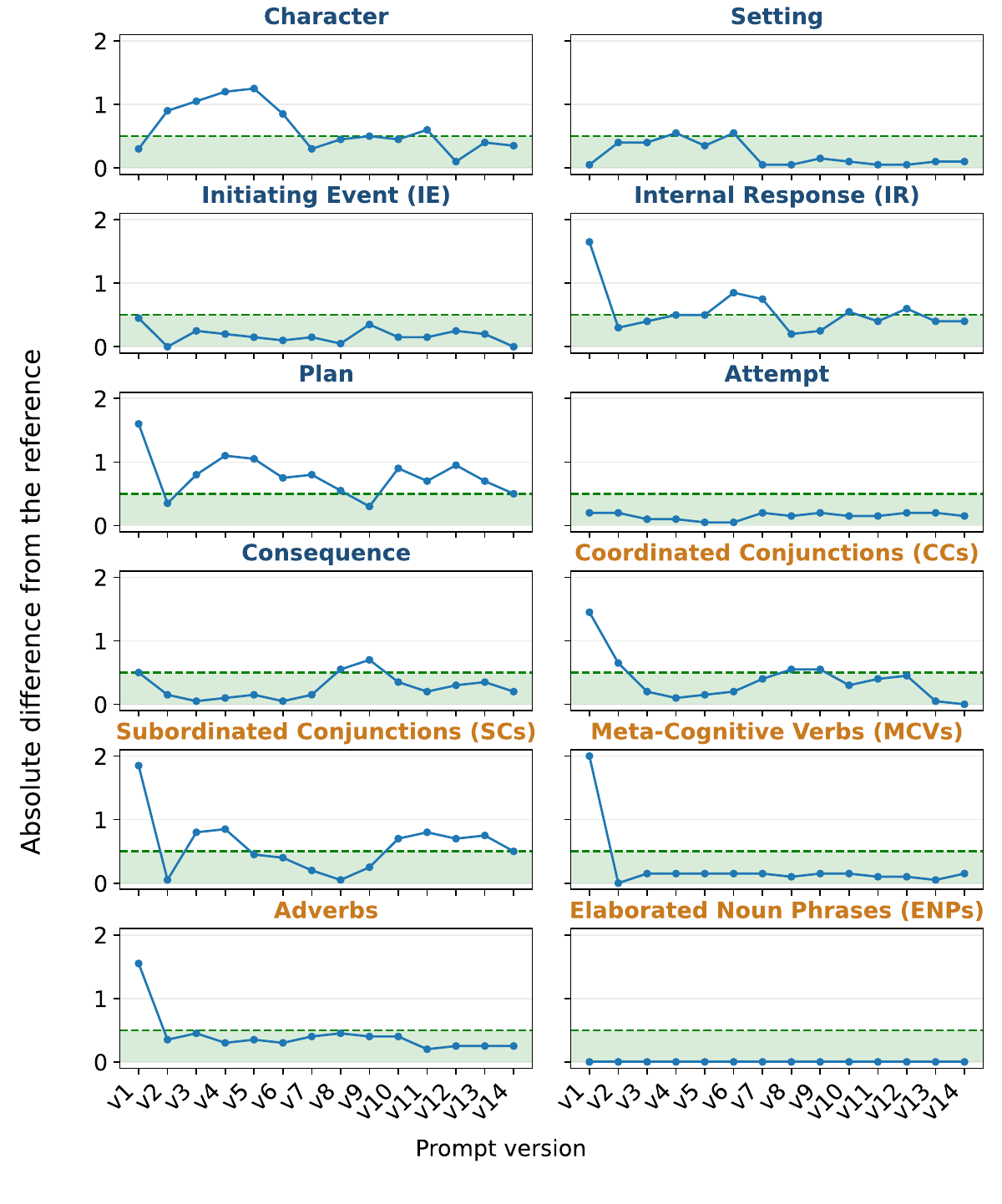}
    \vspace{-5mm}
   \caption{\textbf{Prompt calibration across iterative versions.} Each panel reports the average difference between the MISL feature score obtained with a given prompt version and the reference value. Macro (blue titles) and micro-structural (orange titles) items are shown separately. The green shaded area denotes the target calibration region (difference $\leq0.5$).}
   \label{fig:tuning_prompt}
\end{figure}

At each calibration iteration, the candidate prompt was used to generate 20 stories from the same predefined set of three stickers selected by the participant. Keeping the visual stimuli constant allowed us to compare prompt versions under equivalent conditions, reducing the influence of stimulus-related variability. The generated stories were automatically scored according to the twelve MISL dimensions. For each feature, the mean score was compared with the corresponding target value derived from the human-annotated reference stories. The observed feature-level discrepancies guided the subsequent revision of the prompt. For example, insufficient scores for Internal Response or Plan resulted in more explicit instructions to introduce characters' emotions or future actions. The calibration ended when the absolute difference between the generated and target mean scores was no greater than 0.5 points for every MISL dimension, corresponding to half of one scoring level, as illustrated in Fig.\ref{fig:tuning_prompt}.
Although prompt revision was manually performed in this study, the procedure establishes the basis for a closed-loop calibration architecture in which discrepancies between the desired and observed MISL profiles are automatically used to adapt the generation instructions to specific narrative requirements (Fig. \ref{fig:agents}B). 

\subsection{Painter agent: Visual story representation}
After each narrative turn, the story was illustrated through a two-stage pipeline to support shared attention and narrative continuity across human and robot turns (Fig.~\ref{fig:setup}B). A \emph{Visual Prompt Designer} (based on GPT-4.1) combined the current narrative segment with the preceding visual context to generate an image prompt specifying relevant characters and settings while preserving visual consistency across turns. An \emph{Image Generator} (based on GPT Image 1.5) then produced a child-appropriate illustration, displayed to both the participant and the robot as part of the shared narrative context (see Fig. \ref{fig:agents}C).

\section{METHODS}

\subsection{Participants}
Forty healthy and neurotypical participants were recruited in the experiment (mean age = 36.05 ± 12.73 yrs, 28 females). Three participants were excluded from the analyses due to technical issues with microphone recordings and/or speech transcription during data acquisition. None of the participants had direct experience in the robotics field and gave written informed consent to participate in the study. Data collection was carried out in compliance with the Ethical standards of the 2013 Declaration of Helsinki, and the procedures received approval from the regional ethics committee (Comitato Etico Regione Liguria). All participants received an honorarium of 15 euros for participating.

\subsection{Experimental procedure}

A within-subject design was adopted, with each participant interacting with the robot under two experimental conditions: ``Cond. A- Simple prompt'' and ``Cond. B- Complex prompt''. Cond. A employed the prompt from our previous work \cite{lombardi2025would}, whereas Cond. B used the prompt developed through the calibration procedure described in Sec. \ref{sec: prompt-tuning}. The order of conditions was counterbalanced across participants to minimize potential order effects.

At the beginning of each experimental session participants completed two computerized memory tasks (see Section \ref{sec:cognitive}). 
Participants then completed 4 collaborative trials in each condition, for a total of 8 stories. Story generation was facilitated by cubes with stickers (see Fig.~\ref{fig:setup}-B). Each trial consisted of three steps. First, the participant passed a chosen sticker to iCub, which grasped it and, upon detecting the image, generated the intro of the story. Second, the participant advanced the plot verbally, while offering a second sticker, receiving feedback from the robot about their contribution. In the final phase, following the sticker exchange, the robot generated the end of the story. The participant's contribution in the second phase was transcribed and stored together with the generated narrative and system logs. 

Throughout the experiment, participants were asked to answer questionnaires (detailed in Section \ref{sec:ques}). The complete procedure (see Fig.~\ref{fig:setup}-C) lasted $\sim$ 90 minutes.

\subsection{Cognitive tasks}
\label{sec:cognitive}
To explore the relationship between participants' task contributions and cognitive functions associated with narrative abilities, such as memory \cite{malcorra2021low,kormos2011working}, participants completed two computerized memory tasks. Verbal short-term memory (STM) was assessed with a letter span task, in which participants recalled letter sequences until performance fell below criterion, yielding the individual span. Non-verbal working memory (WM) was assessed with a visual WM task requiring participants to identify whether a probe item had appeared in an array that had been earlier presented briefly. The array could contain varying numbers of items with performance measured as the proportion of correct responses.

\subsection{Questionnaires}
\label{sec:ques}
Subjective measures were collected at three stages of the experiment (see Table \ref{tab:adhoc_questionnaires}). 

\paragraph{Trial-level evaluation}
At the end of each trial, participants completed a two-item ad-hoc questionnaire assessing (i) how much they liked the story and (ii) how coherent they perceived it to be, on a 10-point scale.

\paragraph{Post-condition evaluation}
After completing all four trials of each condition, participants filled out the System Usability Scale (SUS)~\cite{brooke1996sus} and an ad-hoc questionnaire. The SUS consists of ten statements rated on a five-point Likert scale, ranging from 1 (\emph{strongly disagree}) to 5 (\emph{strongly agree}). Positively and negatively worded items are scored according to the standard SUS procedure, producing an overall usability score ranging from 0 to 100.
The ad-hoc questionnaire comprised seven items investigating participants' experience with iCub \cite{lombardi2025would}.

\paragraph{Post-experiment comparison}
At the end of the experiment, participants were asked three additional questions to directly compare their experiences in the two Conditions.

\begin{table*}[t]
\centering
\caption{\textbf{Items included in the ad-hoc questionnaires administered at different stages of the experiment.}}
\vspace{-3mm}
\label{tab:adhoc_questionnaires}

\scriptsize
\setlength{\tabcolsep}{4pt}
\renewcommand{\arraystretch}{0.95}

\begin{tabularx}{\textwidth}{
    >{\centering\arraybackslash}p{0.035\textwidth}
    >{\raggedright\arraybackslash}X
    >{\centering\arraybackslash}p{0.035\textwidth}
    >{\raggedright\arraybackslash}X
}
\toprule

\rowcolor{gray!20}
\multicolumn{4}{l}{\textbf{Trial-level evaluation}} \\
\addlinespace[1pt]
1 & How much did you enjoy the story? Please rate it from 1
(``I did not enjoy it at all'') to 10
(``I enjoyed it very much''). &
2 & How coherent did you find the story? Please rate it from 1
(``The story did not make sense at all'') to 10
(``The story was very coherent and flowed smoothly''). \\
\midrule

\rowcolor{gray!20}
\multicolumn{4}{l}{\textbf{Post-condition evaluation}} \\
\addlinespace[1pt]
1 &
How comfortable did you feel while interacting with the robot? &
5 &
How much did you enjoy interacting with iCub? \\

2 &
Was the number of available stickers sufficient to create an engaging
story? If not, how many stickers would you have preferred? &
6 &
Would you change anything about this activity with the robot? If yes, what? \\

3 &
Was the number of storytelling turns sufficient to create an engaging
story? If not, how many turns would you have preferred? &
7 &
Would you have liked to continue interacting with iCub? \\

4 &
Which groups of people or age ranges do you think this robotic
application would be suitable for? &
& \\

\midrule

\rowcolor{gray!20}
\multicolumn{4}{l}{
    \textbf{Post-experiment comparison}
} \\
\addlinespace[1pt]

1 &
Did you prefer one of the two storytelling sessions with iCub?
If so, which one and why? &
2 &
Did you feel that you contributed more to the stories in one session
than in the other? If so, which one? \\

3 &
Did the images generated and displayed on the screen help you contribute
to the stories? &
& \\
\bottomrule
\end{tabularx}
\end{table*}

\subsection{Measures}

Human narrative production was evaluated at three complementary levels.

\paragraph{Verbal productivity}
We computed the number of words in the human-generated continuation of each story.

\paragraph{Narrative microstructure}
The human continuation was scored independently using the five MISL microstructure dimensions. This analysis evaluates the linguistic structures explicitly produced by the participant without including language generated by the robot.

\paragraph{Narrative-trajectory alignment}
MISL macrostructure depends on causal and temporal relations spanning the evolving narrative and therefore cannot be interpreted reliably from an isolated turn. To investigate whether participants followed the narrative trajectory established by the artificial partner, we generated AI-only counterfactuals for each trial by replacing the participant's second turn with AI-generated continuations (Cond. A $\&$ B) while preserving the same narrative context and visual stimulus. We then compared the cumulative macrostructure score after the second turn for the human-in-the-loop and AI-only versions. Rather than measuring participants' independent storytelling ability, this analysis evaluates whether the joint story reaches a macrostructural complexity comparable to that produced autonomously by the AI from the same opening context.

\section{RESULTS}

\subsection{\textbf{RQ1: Are human verbal productivity and linguistic narrative complexity influenced by the narrative strategy adopted by the humanoid robot?}}

\subsubsection*{Verbal productivity}

Participants produced longer continuations in Cond. B. The mean human contribution contained 20.71 words ($SD=13.01$) in Cond. A and 24.14 words ($SD=13.29$) in Cond. B. A paired one-tailed t-test confirmed a significant increase in verbal production, $t(36)=2.505$, $p=.008$. At participant level, $70.3\%$ of users produced a larger mean number of words in Cond. B. 

Although participants produced more words in absolute terms in Cond. B, their contribution represented a smaller proportion of the complete story. The completed stories contained, on average, 46.78 words ($SD=15.07$) in Cond. A and 104.86 words ($SD=17.48$) in Cond. B. Consequently, the human contribution represented approximately 44.3\% in Cond. A and 23.0\% in Cond. B stories. Thus, the two measures capture different aspects of contribution: participants produced more words in Cond. B, but the robot accounted for a larger proportion of the entire story.

\subsubsection*{Narrative microstructure}

The human-generated continuations obtained a mean micro-MISL score of 4.05 ($SD=1.88$) in Cond. A and 4.91 ($SD=2.11$) in Cond. B. The increase of 0.86 points corresponds to 5.7\% of the complete micro-MISL scale range and to a 21.2\% increase relative to Cond. A mean. The paired one-tailed t-test indicated that the human continuations produced in Cond. B had significantly richer microstructure, $t(36)=3.83$, $p<.001$. Exploratory feature-level analyses indicated greater use of subordinating conjunctions, meta-cognitive verbs, and adverbs in Cond. B. Therefore, the longer continuations elicited in Cond. B were also accompanied by higher linguistic microstructure scores.

\subsubsection*{Overall complexity of the collaborative narratives}

As a complementary system-level measure, we evaluated the global MISL score of the completed human--robot stories. The narratives obtained a mean score of 15.56 ($SD=2.03$) in Cond.~A and 25.14 ($SD=1.81$) in Cond.~B. This difference was statistically significant, $t(36)=-25.8$, $p<.001$, indicating that the experimentally manipulated robot behavior remained observable in the final joint outcome.

However, this measure includes both the opening and the final turn generated by the robot. It should therefore be interpreted as an outcome of the complete human--robot system rather than as a direct measure of the participant's contribution. The participant's role in the development of narrative macrostructure is examined in RQ2 through an AI-only counterfactual comparison.

\subsection{\textbf{RQ2: How does the narrative developed with the human compare with the complexity that the robot would autonomously produce from the same interaction context?}}

To contextualize the human contribution, we generated AI-only continuations for each trial. The counterfactual pipeline received the same AI-generated opening and the same participant-selected stimulus used in the corresponding trial, but the second narrative turn was generated by the AI.

\begin{figure}[h]
   \centering
    \includegraphics[width=\columnwidth]{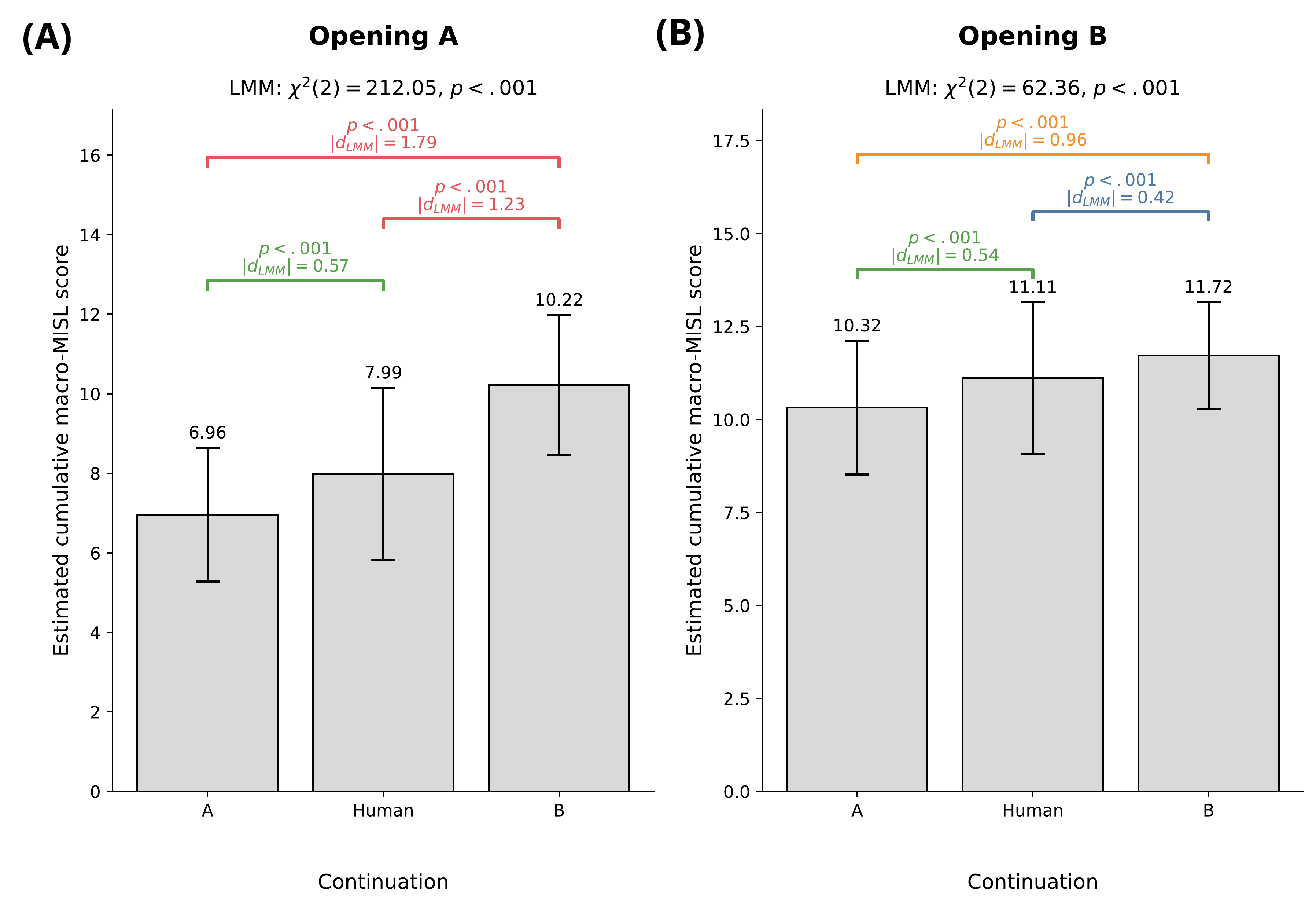}
   \caption{\textbf{Cumulative macro-MISL scores for each continuation type under the two opening conditions.} Bars correspond to the means, whereas error bars represent the standard deviation of the scores. Pairwise contrasts are annotated with Holm-adjusted p-values and residual-standardized effect sizes ($d_{LMM}$); bracket colors indicate effect-size magnitude according to conventional thresholds: small (blue, $d_{LMM}<0.50$), medium (green, $0.50\leq d_{LMM} <0.80$), large (orange, $0.80\leq d_{LMM} <1.20$), and very large (red, $d_{LMM}\geq1.20$).}
   \label{fig:lmm}
\end{figure}

Because MISL macrostructure depends on causal and temporal relations extending across narrative turns, we compared the cumulative macro-score after the second narrative turn. For the same-opening comparison, human-in-the-loop narratives obtained a mean macro-MISL score of 8.00 versus 7.00 for the corresponding AI-only counterfactuals in Cond. A, whereas in Cond. B the respective scores were 11.10 and 11.70. Linear mixed-effects models revealed a significant effect of continuation type in both opening conditions (Cond. A: $\chi^2 (2)=212.05$, $p<.001$; Cond. B: $\chi^2(2)=62.36$, $p<.001$). Pairwise comparisons indicated that the human continuations differed significantly from the corresponding AI-generated continuations in both conditions (Cond. A: $d_{LMM}=0.57$; Cond. B: $d_{LMM}=0.42$), although these effects were moderate and small, respectively (see Fig. \ref{fig:lmm}).

When compared with the AI continuations generated under the alternative opening condition, substantially larger differences emerged. In Cond. A, the human narratives differed from the alternative AI continuation by 2.23 macro-MISL points ($d_{LMM}=1.23$), while the direct comparison between the two AI prompting strategies yielded a difference of 3.26 points ($d_{LMM}=1.80$). Similarly, in Cond. B, the human narratives differed from the alternative AI continuation by 0.79 points ($d_{LMM}=0.54$), whereas the two AI prompting strategies differed by 1.40 points ($d_{LMM}=0.96$). All pairwise comparisons were statistically significant after Holm correction ($p<.001$). Overall, the comparison with the corresponding AI counterfactual consistently yielded smaller discrepancies than the comparison with the alternative prompting strategy. These findings suggest that participants generally developed their narratives within the macrostructural trajectory established by the robot they interacted with, while remaining clearly distinguishable from narratives generated under a different prompting strategy.

\subsection{\textbf{RQ3: Does a more complex storytelling strategy improve perceived story quality and interaction preference without negatively affecting usability or engagement?}}

\subsubsection*{Trial-level evaluation} Participants rated stories in Cond.~B as significantly more likeable and coherent than those in Cond.~A
(see Table~\ref{tab:subjective_results}). This pattern was also evident at participant level: $78.4\%$ assigned higher mean likeability ratings to Cond.~B, while $70.3\%$ assigned higher coherence ratings.

\subsubsection*{Post-condition evaluation}
No significant differences emerged in system usability, comfort, or engagement (see Table~\ref{tab:subjective_results}). SUS scores remained high in both conditions, suggesting that the more complex narrative strategy preserved the usability of the interaction. Similarly, participants
reported high comfort and engagement in both conditions.

Other questionnaire responses descriptively favored Cond.~B: more participants wished to continue interacting with iCub ($78.4\%$ vs.\ $70.3\%$) and considered the available stickers ($81.1\%$ vs.\ $70.3\%$) and number of turns ($70.3\%$ vs.\ $56.8\%$) sufficient. Furthermore, fewer participants in Cond. B felt the need to suggest changes to the task ($37.8\%$ vs.\ $45.9\%$). However, these differences were not statistically significant in mixed-effects logistic regression models (all $ps \geq .162$). Participants indicated that approximately five narrative turns would be ideal in both conditions (Cond.~A: $M = 4.88$, $SD = 1.63$; Cond.~B: $M = 5.18$, $SD = 1.66$), consistent with our previous study \cite{lombardi2025would}. The present protocol used three turns to limit the duration of the eight collaborative trials.

\begin{figure}[h]
   \centering
    \includegraphics[width=0.9\columnwidth]{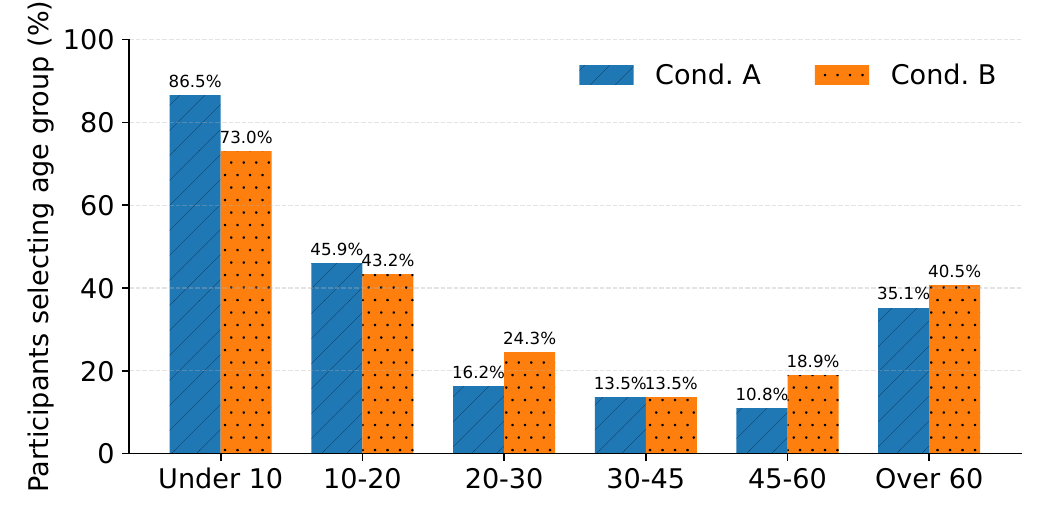}
   \caption{\textbf{Target groups.} Percentage of participants selecting each age group as the most suitable target population for the proposed framework after interacting with iCub in Cond. A and Cond. B- multiple selections were allowed; therefore, percentages do not sum to $100\%$. }
   \label{fig:age_group}
\end{figure}

Regarding potential target populations, participants primarily selected children under 10, followed by adolescents and older adults (Fig.\ref{fig:age_group}). Age-group selections differed significantly within both conditions (Cochran's $Q$: Cond.~A, $Q(5) = 75.44$; Cond.~B, $Q(5) = 42.93$; both $ps < .001$).

\subsubsection*{Post-experiment comparison}
A direct preference was expressed by $78.4\%$ of participants; of these, 25 ($86.2\%$) preferred Cond.~B. However, only $48.6\%$ perceived a difference in their own contribution between conditions, and $55.6\%$ of this subgroup reported contributing more in Cond.~B. Thus, the objective increases in word production and microstructure were not accompanied by a strong conscious perception of increased contribution. Finally, $62.2\%$ considered the generated images helpful for memorizing, visualizing, and/or developing the story.

\begin{table}[t]
\centering
\caption{\textbf{Subjective evaluations under
the two narrative conditions.}}
\vspace{-3mm}
\label{tab:subjective_results}

\scriptsize
\setlength{\tabcolsep}{3pt}
\renewcommand{\arraystretch}{0.95}

\begin{tabularx}{\columnwidth}{
    >{\raggedright\arraybackslash}X
    >{\centering\arraybackslash}c
    >{\centering\arraybackslash}c
    >{\centering\arraybackslash}c
    >{\centering\arraybackslash}c
}
\toprule
\textbf{Measure} &
\textbf{Cond. A} &
\textbf{Cond. B} &
\textbf{$t(df)$} &
\textbf{$p$} \\
\midrule

\rowcolor{gray!20}
\multicolumn{5}{l}{\textbf{Trial-level evaluation}} \\
\addlinespace[1pt]

Likeability &
7.30 $\pm$ 1.95 &
8.08 $\pm$ 1.50 &
$3.69(36)$ &
$<.001$ \\

Coherence &
7.17 $\pm$ 1.93 &
7.89 $\pm$ 1.55 &
$3.49(36)$ &
$<.001$ \\

\midrule
\rowcolor{gray!20}
\multicolumn{5}{l}{\textbf{Post-condition evaluation}} \\
\addlinespace[1pt]

Usability (SUS) &
77.43 $\pm$ 14.31 &
78.24 $\pm$ 12.19 &
$0.85(36)$ &
$.399$ \\

Comfort &
8.40 $\pm$ 1.46 &
8.43 $\pm$ 1.54 &
$0.16(36)$ &
$.872$ \\

Engagement &
8.59 $\pm$ 1.86 &
8.73 $\pm$ 1.57 &
$0.93(36)$ &
$.360$ \\

\bottomrule
\end{tabularx}

\vspace{1mm}
\parbox{\columnwidth}{\scriptsize
Values are means $\pm$ standard deviations. Ratings range from 1 to 10, except SUS (0--100). Statistics refer to paired-samples $t$-tests.}
\end{table}

\subsection{\textbf{RQ4: Are individual cognitive abilities associated with collaborative storytelling performance?}}

We investigated whether participants' cognitive abilities influenced their narrative production in this task. Specifically, we analyzed the relationship between memory test scores and the objective measures of narrative production using linear mixed-effects models (LMMs). Five additional participants were excluded from this analysis due to task comprehension issues or to poor performance (accuracy below $50\%$ in the visual WM task), leaving a final sample of 32 participants. For verbal productivity, verbal STM capacity (letter span task) exhibited a marginal positive trend ($F(1, 29) = 3.45$, $p = .073$), while non-verbal WM accuracy remained non-significant ($F(1, 29) = 0.04$, $p = .843$). Similarly, for narrative microstructure, verbal STM capacity demonstrated a marginal significant trend ($F(1, 29) = 3.79$, $p = .061$), whereas visual WM accuracy showed no significant association ($F(1, 29) = 0.001$, $p = .971$). For narrative-trajectory alignment- macrostructure-, neither verbal STM capacity ($p = .634$) nor non-verbal WM ($p = .692$) predicted alignment outcomes. Crucially, neither memory metric interacted with task condition ($ps \geq .382$), showing that individual memory capacities did not differentially modulate narrative performance across complexity levels (controlling for Condition, $ps \leq .019$). These findings provide no clear evidence that the measured memory capacities account for the observed differences in narrative performance, which appear more closely associated with the experimental storytelling conditions.

\section{CONCLUSIONS}
This work investigated whether the narrative strategy adopted by an LLM-driven humanoid robot influences human production during collaborative storytelling. To address this question, we extended the previous storytelling framework \cite{lombardi2025would} with an automatic MISL-based narrative analyzer, a human-in-the-loop prompt calibration procedure, and a visual story generation pipeline.

The experimental results indicate that the narrative behavior of the robot affects the way participants contribute to the shared story.
When interacting with a more complex storytelling partner, participants produced longer narrative continuations and achieved higher micro-MISL scores, suggesting that richer robot contributions elicited richer linguistic production. At the same time, stories generated in the complex condition were perceived as more enjoyable and coherent and were preferred by most participants, while usability, comfort, and engagement remained comparable.

The AI-only counterfactual analysis further suggested that the macrostructural evolution of the jointly generated stories remained close to the trajectories autonomously produced by the robot, with only small absolute differences between the two conditions.
Although statistical equivalence was not established, these results indicate that the human contribution remained within a comparable macrostructural range while exhibiting different adaptation patterns depending on the narrative strategy adopted by the robot.

Taken together, these findings suggest that narrative complexity should be regarded not only as a property of generated stories, but also as a tunable interaction parameter capable of modulating human participation and alignment during collaborative storytelling. Rather than always maximizing narrative richness, future socially assistive robots could adapt their storytelling strategy according to the user's ongoing performance, alternating between richer narrative scaffolds and more minimal prompts to encourage active participation.

Future work will investigate adaptive closed-loop storytelling architectures in which the automatic analyzer continuously monitors the user's narrative production and dynamically adjusts the robot's storytelling strategy in real time. Additional studies involving more interaction turns, larger and more diverse participant populations, and longitudinal protocols will be necessary to determine whether the short-term modulation observed in this work translates into sustained linguistic and cognitive benefits. Such adaptive narrative human--robot interaction paradigms may ultimately support personalized tools for cognitive stimulation, assessment, and rehabilitation.

\addtolength{\textheight}{-12cm}   





\bibliographystyle{IEEEtran}
\bibliography{biblio}

\end{document}